\documentclass[10pt,twocolumn,letterpaper]{article}

\usepackage[pagenumbers]{cvpr} 
\usepackage{amsmath,graphicx}
\usepackage{adjustbox}
\usepackage{url}
\usepackage{array}
\usepackage{multirow}
\usepackage{amssymb}
\usepackage{pifont}
\usepackage{pgfplots}
\usepackage{float} 
\usepackage{multirow}
\usepackage{booktabs}
\usepackage{makecell}
\usepackage{tikz}
\usepackage{tipa}
\usepackage{setspace}
\usepackage{xcolor}
\usepackage{caption}
\usepackage{subcaption}
\usepackage{lipsum}
\usepackage[table]{xcolor}
\usepackage{arydshln}
\usepackage{CJKutf8}
\newcommand{\cmark}{\ding{51}}
\newcommand{\xmark}{\ding{55}}

\definecolor{lightblue}{rgb}{0.93,0.95,1.0} 

\usepackage{bbm} 
\definecolor{cvprblue}{rgb}{0.21,0.49,0.74}
\usepackage[pagebackref,breaklinks,colorlinks,allcolors=cvprblue]{hyperref}

\def\paperID{11076} 
\def\confName{CVPR}
\def\confYear{2026}

\title{Learning What to Attend First: Modality-Importance-Guided \\ Reasoning for Reliable Multimodal Emotion Understanding}

\author{Hyeongseop Rha\thanks{Equal contribution},
Jeong Hun Yeo\footnotemark[1],
Junil Won,
Se Jin Park,
and Yong Man Ro\thanks{Corresponding author}\\
Integrated Vision and Language Lab, KAIST, South Korea\\
291 Daehak-ro, Yuseong-gu, Daejeon 34141, Republic of Korea\\
{\tt\small \{ryool\_1832, sedne246, dnjswnsdlf48, jinny960812, ymro\}@kaist.ac.kr}
}

\begin{document}
\maketitle
\begin{abstract}
In this paper, we present Modality-Importance-Guided Reasoning (MIGR), a framework designed to improve the reliability of reasoning-based multimodal emotion understanding in multimodal large language models. Although existing methods have advanced emotion understanding, they often suffer from reasoning drift: models gradually rely on their own generated text instead of multimodal evidence, and their explanations are overly shaped by visually initiated reasoning paths. To address these issues, we introduce Modality Importance (MI), a simple yet effective mechanism for identifying the emotion-dominant modality. Using MI, MIGR reorganizes reasoning sequences so that explanations begin from the modality most critical to the target emotion, preventing early reasoning from being misled by less informative cues. Our two-stage framework—comprising modality-aligned supervised fine-tuning and modality-aware reward optimization—encourages models to generate emotionally grounded, causally relevant, and coherence-preserving explanations. Experimental results on the DFEW benchmark show that MIGR substantially improves reasoning reliability, decreasing instances of correct predictions accompanied by emotionally inconsistent explanations from 18.10\% to 7.37\%. These results confirm the benefit of initiating reasoning from the emotion-dominant modality.
\end{abstract}

\section{Introduction}
\label{sec:intro}
As human–machine interaction becomes increasingly prevalent through AI assistants, smart speakers, and wearable devices, understanding human emotions has become crucial for achieving empathetic and socially aware communication. To address this need, Multimodal Emotion Recognition (MER)~\cite{kossaifi2017afew,zadeh2018multimodal,poria2018meld,busso2008iemocap,perepelkina2018ramas,liu2022mafw,jiang2020dfew,livingstone2018RAVDESS} has emerged as a key research direction, leveraging complementary cues from visual, acoustic, and textual modalities to infer emotional states more accurately than unimodal approaches~\cite{shou2025mer_survey}. Recent progress in multimodal integration with Large Language Models (\ie, MLLMs) has significantly improved MER performance, enabling models to more effectively leverage the complementary cues~\cite{xing2024emo,cheng2024emotion,yang2025omni,hu2025feallm, zhao2025favchat}. Building on this foundation, the introduction of Reinforcement Learning (RL) optimization has further extended MER from recognition to reasoning-based emotion understanding, allowing models to articulate why an emotion arises based on multimodal evidence~\cite{zhao2025r1}.

\begin{figure}[t]
    \centering
    \includegraphics[width=0.48\textwidth]{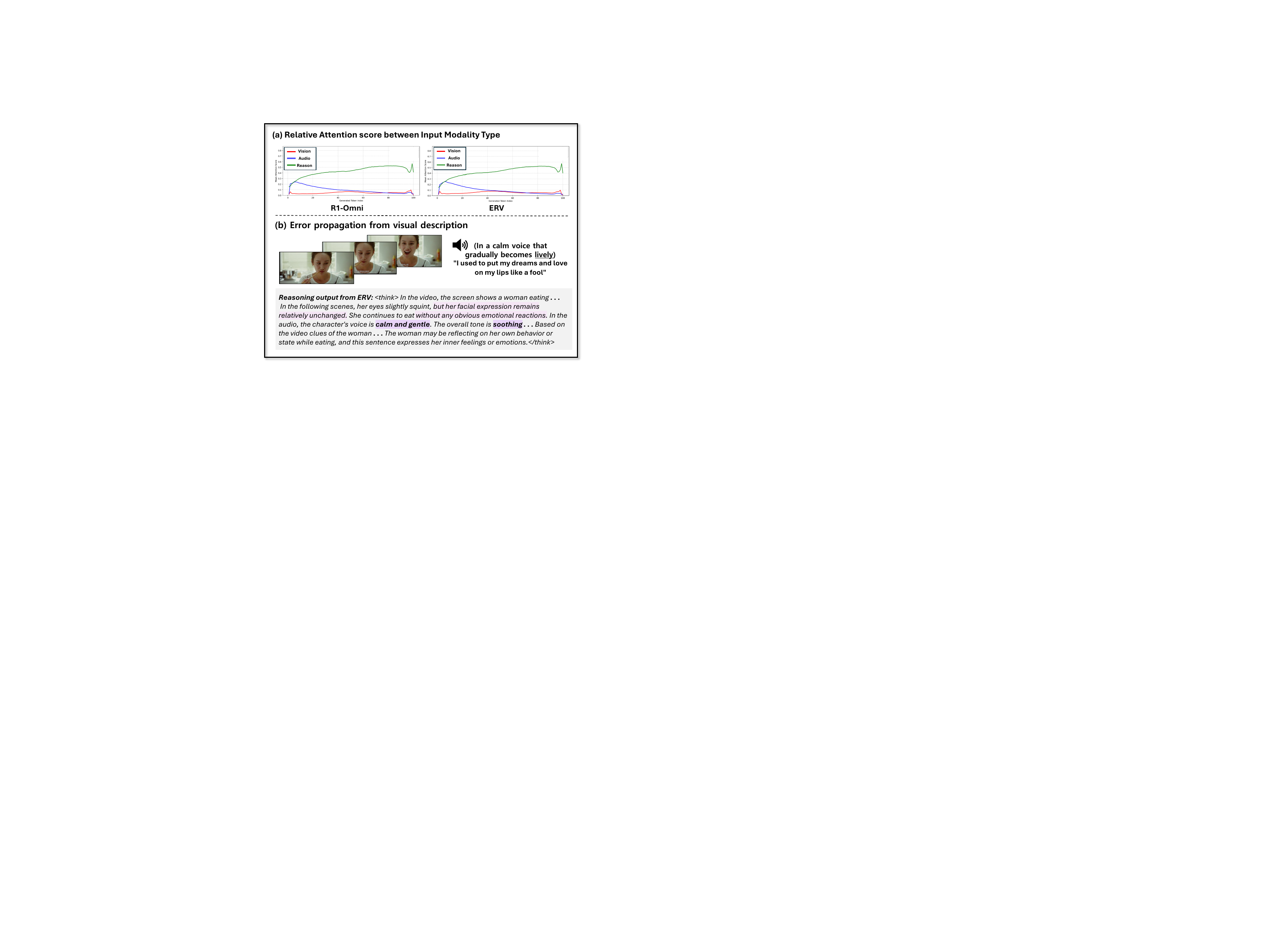}
    \caption{Failure pattern of the baseline multimodal reasoning model. (a) The model’s attention gradually drifts from external multimodal inputs to previously generated text. (b) Due to the fixed training order: visual description → audio description → reasoning, the model often begins with a visual-based summary; when this initial step is inaccurate or emotionally irrelevant, subsequent reasoning becomes increasingly misaligned.}
    \label{fig:1}
    \vspace{-0.2cm}
\end{figure}

Despite the recent progress, reasoning-based emotion understanding with MLLMs remains relatively underexplored. Our aim is to achieve reliable emotional reasoning, which we define as the ability to generate explanations that are consistent with the model’s final emotion prediction. It is an essential capability for emotional AI, which must not only predict emotions accurately but also justify them with consistent and trustworthy explanations in ambiguous multimodal scenarios. However, to the best of our knowledge, only one prior work has attempted to improve such reliability~\cite{rha2025erv}. It introduces RL optimization that rewards MLLMs when an LLM judges the generated reasoning to be consistent with the prediction. While this approach shows some improvements, it remains an early step toward reliable emotional reasoning, and substantial further advances are still required.

In this context, we begin by exploring where this recent attempt~\cite{rha2025erv} falls short. As shown in Figure~\ref{fig:1}(a), the model’s attention gradually shifts from multimodal inputs (\ie, audio and visual cues) to its own generated text as reasoning progresses. This shift causes the model to anchor its subsequent reasoning on whatever emotional cues appear in the initial generated text. Moreover, the reasoning process typically starts from generating a video description, because the training data are organized in a fixed order in which visual descriptions precede audio descriptions and the final reasoning. However, such a visual-first reasoning path may not reflect the true emotional driver, especially in cases where emotion is conveyed primarily through vocal tone or semantics. When the initial visual-based reasoning is irrelevant or inaccurate, the subsequent reasoning becomes increasingly misaligned, as illustrated in Figure~\ref{fig:1}(b). As a result, the initial step of reasoning becomes disproportionately influential. Similar issues appear in other domains, where early cues steer reasoning away from the underlying evidence~\cite{luo2025thinking}.

To address the issue, we propose Modality-Importance-Guided Reasoning (MIGR), a method that aims to initiate reasoning from the emotion-dominant modality (\ie, the modality most strongly associated with the target emotion) and maintain emotion-coherent reasoning throughout inference. To achieve this, we introduce Modality Importance (MI) estimation that identifies the modality most critical for accurate emotion recognition by comparing the model’s predicted emotion under audio-only, visual-only, and audio–visual inputs. If the model accurately predicts the emotion under audio-only input but fails under visual-only or audio–visual conditions, this suggests that the visual modality may be irrelevant or even misleading. In this case, the audio modality is considered emotion-dominant.

Building on this MI and recent reasoning-based MLLM training paradigms~\cite{zhao2025r1, rha2025erv}, we design our method as a two-stage learning framework consisting of (i) modality-aligned Supervised Fine-Tuning (SFT) and (ii) modality-aligned reward optimization. In the modality-aligned SFT stage, we leverage the MI to reorganize the original reasoning text into modality-specific segments and then reorder according to emotion-dominant modality. Then, we train on the data so that it starts to generate emotion-relevant reasoning consistently at the beginning. In the modality-aligned reward optimization, we depart from conventional RL reward designs~\cite{zhao2025r1} that primarily focus on reasoning format or answer accuracy, and instead enforce modality-aware constraints through two complementary rewards. The modality-aligned order reward encourages the model to initiate its reasoning from the emotion-dominant modality, thereby reinforcing a modality-aligned reasoning structure. The modality-grounded reasoning reward further promotes generating emotion-dominant modality reasoning text that is causally and semantically aligned with the target emotion.

In this work, we make the following contributions: (i) We introduce the MI for identifying the emotion-dominant modality in multimodal emotion reasoning. It provides a simple yet effective way to diagnose misleading and informative modalities for modality-aligned emotion reasoning. (ii) We propose MIGR which restructures reasoning data so that the model initiates reasoning from the most informative modality. MIGR further reinforces this modality-aligned reasoning through modality-aware reward optimization, producing more stable and emotion-grounded explanations. (iii) MIGR achieves substantial improvements in reasoning reliability on the DEFW benchmark, including a 15.42\% gain in Explanation–Prediction Consistency and a reduction of emotionally incorrect reasoning from 18.10\% to 7.37\%. These gains demonstrate the effectiveness of emotion-dominant modality alignment for robust multimodal emotional reasoning.

\section{Related Work}
\subsection{Multimodal Emotion Recognition}
\subsubsection{Traditional MER Approaches}
MER aims to understand human affective states by analyzing multimodal signals that appear across diverse expressive scenarios. The development of MER was initially driven by the introduction of unimodal emotion-related datasets~\cite{busso2008iemocap,livingstone2018RAVDESS,kossaifi2017afew,jiang2020dfew,wang2022ferv39k}, which provided isolated affective cues such as audio, visual, or speech signals. Leveraging these datasets enabled researchers~\cite{zhang2017speech,zhang2021learning, zhao2021former,sun2023mae,chumachenko2024mma}  to quantitatively analyze emotional expressions within each modality, leading to the development of modality-specific encoders that effectively model and encode unimodal emotional information. Subsequently, the introduction of multimodal emotion datasets~\cite{zadeh2018multimodal,poria2018meld,perepelkina2018ramas,liu2022mafw,lee2019CAER} further accelerated the progress of MER by offering richer combinations of affective cues—including audio, visual, textual, and physiological signals. These datasets opened the door for modeling cross-modal interactions, driving the development of fusion strategies~\cite{narayan2024facexformer,sun2024hicmae} designed to integrate complementary information across modalities and improve recognition robustness.

\subsubsection{Multimodal LLM-based MER}
With the recent success of MLLMs across a wide range of domains, these models have also begun to be applied to emotion understanding tasks. The introduction of multimodal emotion-descriptive datasets such as EMER~\cite{lian2023EMER} has further accelerated progress in MLLM-based MER, enabling MLLMs to approach emotion recognition as a natural language–based reasoning and explanation task~\cite{xing2024emo,cheng2024emotion,lian2025affectgpt,yang2025omni,hu2025feallm, zhao2025favchat}. However, such emotion descriptions require manual annotation, making large-scale expansion difficult. To address this limitation, recent studies~\cite{cheng2024emotion,lian2024open,yang2025omni} have proposed leveraging MLLMs to automatically generate emotion descriptions as pseudo-labels, enabling scalable construction of description-enriched datasets. Building upon these expanded datasets, several works~\cite{cheng2024emotion,lian2024open} have explored generating and interpreting emotion explanations grounded in video and audio cues. In parallel, architectural advances~\cite{zhao2025humanomni,lian2025affectgpt} have further improved the accuracy of emotion recognition by enhancing multimodal representation learning and reasoning capabilities.

\subsubsection{Reasoning with Multimodal LLMs}
Despite significant progress in MLLM-based MER, these approaches still rely heavily on well-designed emotional annotations. This dependency has motivated growing interest in training paradigms that reduce reliance on explicit labels and instead strengthen reasoning through alternative supervision. In this context, RL-based post-training has recently emerged as an effective strategy for enhancing MLLMs’ reasoning abilities while mitigating label dependence. In particular, GRPO~\cite{shao2024deepseekmath} and Verifiable Reward~\cite{guo2025deepseek} have demonstrated strong generalization not only in math and coding but also across a wide range of multimodal domains~\cite{zhao2025r1,liu2025visualrft,shen2025vlm,park2025dip,chen2025r1v,li2025videochat,lee2025refocus,xing2025echoink}. However, many existing methods primarily focus on optimizing the final answer, while the fidelity and consistency of the intermediate reasoning process often receive insufficient attention, an issue highlighted by~\cite{wei2025gtr}. To address this, recent studies~\cite{chen2025grpocare,yang2025humanomniv2} have introduced consistency-aware methodologies designed to better align reasoning with final outputs. Yet, such reasoning consistency problems remain largely unexplored in inherently ambiguous domains such as MER. Because emotional interpretation depends on subtle and context-dependent multimodal cues, reliable emotional reasoning becomes especially important. In particular, the generated explanations must faithfully support the model’s final emotion prediction. Nevertheless, only one prior work has attempted to improve such reliability in MER~\cite{rha2025erv}, introducing an RL objective that rewards MLLMs when an LLM judge deems the generated reasoning consistent with the prediction. While this approach represents an important initial step, substantial further progress is required to achieve robust and trustworthy emotional reasoning.

\begin{figure*}[t]
    \centering
    \includegraphics[width=0.99\textwidth]{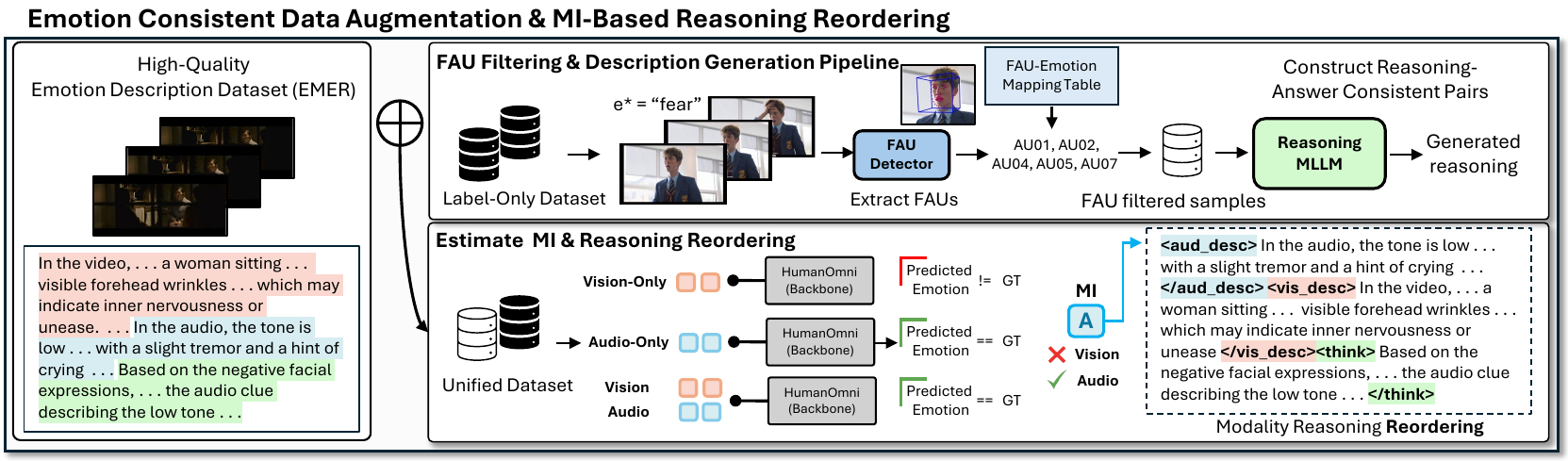}
    \caption{Overview of the proposed data construction pipeline, including FAU-based emotion-consistent data augmentation, MI estimation, and MI-guided modality-specific reasoning reordering.}
    \label{fig:2}
    \vspace{-0.3cm}
\end{figure*}

\section{Method}
In this section, we introduce MIGR, a method that learns which modality to attend to first based on MI. Through this design, we aim to mitigate text biases in the initial reasoning stage and ensure that early reasoning remains grounded in the emotionally dominant modality. Our method consists of two learning stages. First, we present modality-aligned SFT, which enables the model to learn an initial reasoning direction by beginning from the modality identified as emotion-dominant by the MI label. Second, we introduce Modality-Aware Reward Optimization, which aims to reinforce modality prioritization and emotion-grounded reasoning while providing stability during optimization. To this end, we propose two rewards: modality-aligned order reward and modality-grounded reasoning reward. The following sections describe each training stage in detail.

\subsection{Modality-Aligned SFT}
Recent studies~\cite{zhao2025r1} have demonstrated the effectiveness of SFT as a cold-start in emotion understanding tasks, showing that even a small amount of high-quality supervision from the Explainable Multimodal Emotion Reasoning (EMER) dataset~\cite{lian2023EMER} can yield substantial performance gains. Motivated by these findings, we also initiate our training with a cold-start strategy.

\subsubsection{Emotion-Consistent Data Augmentation}
Although EMER~\cite{lian2023EMER} provides high-quality supervision, the SFT stage can be further improved by incorporating samples with particularly clear and consistent emotional cues. To enhance early-stage supervision, we therefore augment our training set with additional emotion-consistent samples. To identify such samples, we extract Facial Action Units (FAUs) from the facial frames of each video, which provide strongly correlated emotional cues~\cite{velusamy2011method}. We then evaluate whether the extracted FAU patterns align with the target emotion based on a predefined FAU–emotion mapping table~\cite{cheng2024emotion}. Only samples whose FAU patterns exactly match the target emotion are selected. For these filtered samples, we construct reasoning–answer pairs using FAU-consistent emotional evidence to guide the reasoning content.

\subsubsection{Modality Importance (MI)}
We now introduce MI that determines the modality most strongly correlated with the target emotional signal. The key idea is to assess the contribution of each modality to the model’s emotional understanding by comparing how accurately the model can infer the target emotion when different modality combinations (i.e., audio-only, visual-only, and audio–visual) are used as inputs. By contrasting the model’s behavior across these modality combinations, we identify which modality is the emotion-dominant in each audio–visual sample.

\subsubsection{MI-Based Modality Reasoning Reordering}
After constructing both the high-quality human-annotated dataset (EMER) and the FAU-filtered additional dataset, we merge them to form a unified training set. Using this combined data, we extract the MI for every sample. Once the emotion-dominant modality is determined, we reorganize the structure of the reasoning text, as shown in Figure~\ref{fig:2}.

Specifically, for each training example, we decompose the reasoning text into two modality-specific reasoning texts: an audio-based reasoning text and a video-based reasoning text. To explicitly mark these modality-specific parts, we employ two special tokens, \texttt{<aud\_desc>} and \texttt{<vis\_desc>}, placing the corresponding token at both the beginning and the end of each audio- and video-based text. We then reorder these text according to the MI. If audio is the emotion-dominant modality, the audio-based reasoning text is placed first, followed by the video-based text; conversely, if video is emotion-dominant, the video-based text is placed before the audio-based text. For samples where audio and video are equally informative, we include both orderings, \ie, one with audio-first and one with video-first. By leveraging the reasoning text reorganized with the dominant modality during SFT, the model learns to initiate its reasoning process from the modality that carries the strongest emotional signal.

\begin{figure*}[h]
    \centering
    \includegraphics[width=0.99\textwidth]{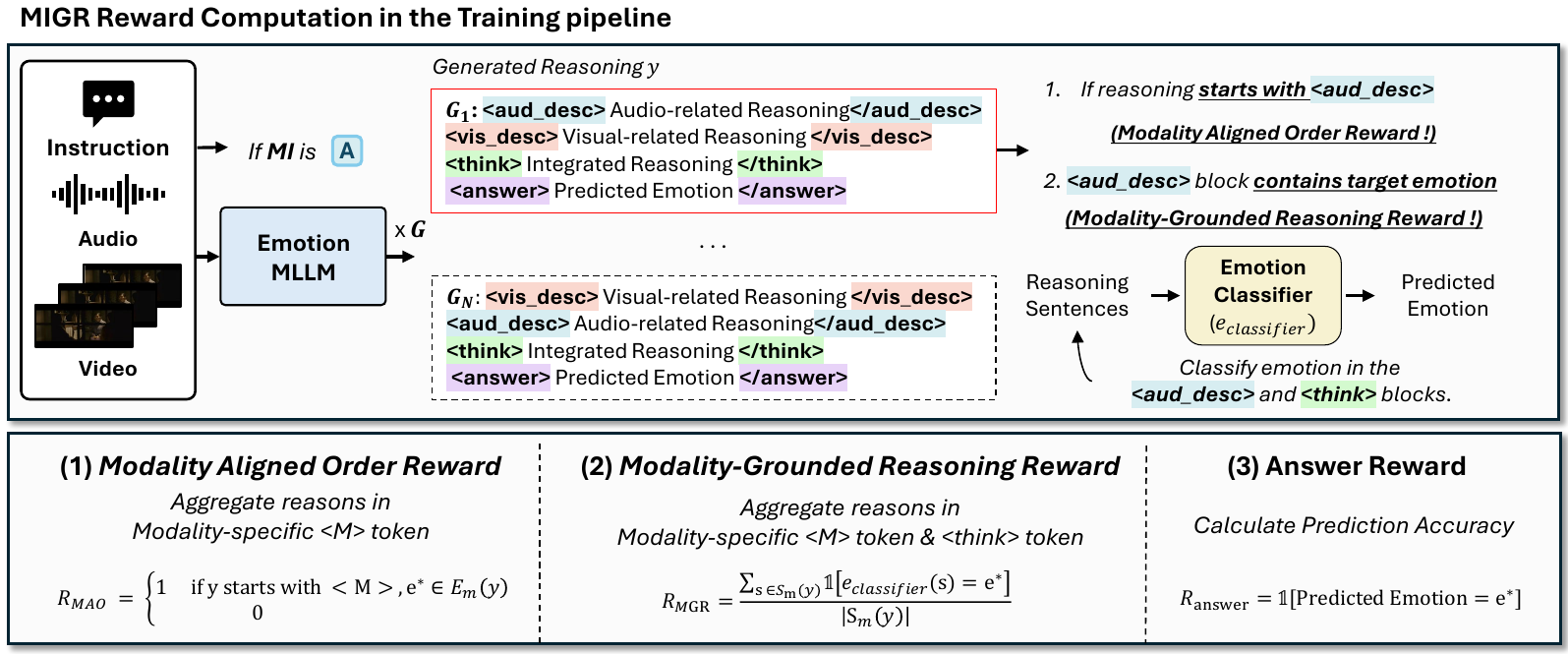}
    \caption{Illustration of the three rewards in MIGR: the Modality-Aligned Order Reward enforces MI-consistent reasoning order; the Modality-Grounded Reasoning Reward ensures emotion-consistent modality-specific reasoning; and the answer reward guarantees correct final emotion prediction.}
    \label{fig:2}
    \vspace{-0.3cm}
\end{figure*}
\subsection{Modality-Aligned Reward Optimization}
Following the cold-start stage, we refine the model’s reasoning process through Modality-Aligned Reward Optimization. Prior GRPO-based approaches~\cite{zhao2025r1, rha2025erv} typically employ two reward types: an answer reward and a format reward. Since these rewards focus only on generating a correct answer and producing well-structured reasoning, they provide no explicit guidance on what kind of reasoning should be produced. Therefore, these two rewards are insufficient to preserve the emotion-dominant modality-prioritized reasoning learned during SFT. To reinforce MI-guided reasoning throughout optimization, we introduce two rewards: the modality-aligned order reward and the modality-grounded reasoning reward.

\subsubsection{Modality-Aligned Order Reward}
The modality-aligned order reward encourages the model to generate reasoning in a modality order consistent with the MI $m$. During optimization, if audio is the emotion-dominant modality, the reward is granted when the model initiates its reasoning with the \texttt{<aud\_desc>} token and the target emotion is detected within the \texttt{<aud\_desc>} token. Similarly, if the video is emotion-dominant, the reward is granted when the reasoning begins with the \texttt{<vis\_desc>} token and the target emotion appears within the \texttt{<vis\_desc>} token.
We denote the modality-specific token corresponding to the MI $m$ as \texttt{<M>}. 
For this token, we define $E_{\text{m}}(y)$ as the set of predicted emotion labels obtained by applying an emotion classifier $e_{classifier}$ to its reasoning sentences.
Under this definition, the modality-aligned order reward is given by:

\begin{equation}
R_{\text{MAO}}(y, m, e^\ast)
=
\begin{cases}
1, & \text{if } y \text{ starts with }\texttt{<M>},\; e^\ast \in E_{\text{m}}(y)
 \\[6pt]
0, & \text{otherwise}.
\end{cases}
\end{equation}

\subsubsection{Modality-Grounded Reasoning Reward}
The modality-grounded reasoning reward evaluates the semantic alignment between each modality-specific reasoning text and the target emotion. For each sample, we measure how well the audio-based and video-based reasoning texts provide emotion-consistent evidence relative to the ground-truth emotion. 
The reward is then computed based on the MI: if audio is the emotion-dominant modality, we aggregate the reasoning sentences from the \texttt{<aud\_desc>} and \texttt{<think>} tokens. Likewise, if video is emotion-dominant, we aggregate the reasoning sentences from the \texttt{<vis\_desc>} and \texttt{<think>} tokens.
An emotion classifier $e_{classifier}$ is then applied to this aggregated set of reasoning sentences, denoted as $S_m(y)$, to obtain the predicted emotion labels.
The correction ratio computed from these predictions is used as the reward.
This reward function can be formulated as follows:

\begin{equation}
R_{\text{MGR}}(y, m, e^\ast)
=
\frac{
\sum_{s \in S_m(y)}
\mathbbm1\left[ e_{\text{classifier}}(s) = e^\ast \right]}{| S_m(y) |}
\end{equation}

In addition to these two modality-focused rewards, we further include an answer reward~\cite{zhao2025r1} ($R_{\text{answer}}$) to ensure the correctness of the final emotion prediction.  In total, MIGR employs three rewards during optimization as follows:

\begin{equation}
R_{\text{total}}
= R_{\text{MAO}} + R_{\text{MGR}} + R_{\text{answer}}.
\end{equation}

As shown in Figure 3, MIGR integrates these three rewards to guide modality-aligned reasoning during optimization.

\section{Experimental Setup}
\subsection{Datasets}
We utilize three multimodal emotion dataset to train our model and validate its effectiveness. \textbf{EMER}~\cite{lian2023EMER} provides high-quality human-verified reasoning–label pairs and is used for cold-start SFT. \textbf{DFEW}~\cite{jiang2020dfew} and \textbf{MAFW}~\cite{liu2022mafw} are large-scale benchmarks covering diverse facial expressions in video, audio, and text modalities.  A detailed description of dataset composition and statistics is provided in the Appendix~\ref{appendix:appendix7.1}.

\subsection{Evaluation Metrics}
Following prior work~\cite{cheng2024emotion, rha2025erv}, we report UAR and WAR for recognition accuracy, and three complementary metrics—EEA, EPC, and FCR—to quantify the emotional coherence of generated reasoning. These metrics jointly assess how well the explanation aligns with the target emotion and the model’s prediction. Formal definitions and annotation protocols are included in the Appendix~\ref{appendix:appendix7.2}.

\begin{table*}[t]
\caption{Comparison with state-of-the-art methods on the MAFW and DFEW datasets. 
We report consistency metrics (FCR, EEA, EPC) and recognition accuracy (UAR, WAR). 
A dagger ($\dagger$) denotes our re-implementation with identical training data for fair comparison.}
\renewcommand{\arraystretch}{1}
\renewcommand{\tabcolsep}{1.8mm}
\centering
\resizebox{0.85\linewidth}{!}{
\begin{tabular}{lcccccccccccc}
\toprule
\multirow{3.5}{*}{\textbf{Model}} 
& \multirow{3.5}{*}{\makecell{\textbf{MLLM} \\ \textbf{Based}}} 
& \multirow{3.5}{*}{\makecell{\textbf{MLLM} \\ \textbf{Params}}} 
& \multicolumn{5}{c}{\textbf{MAFW}} 
& \multicolumn{5}{c}{\textbf{DFEW}} \\
\cmidrule(lr){4-8} \cmidrule(lr){9-13}
& & & \multicolumn{3}{c}{\textbf{Consistency (\%) $\uparrow$}} & \multicolumn{2}{c}{\textbf{Accuracy (\%) $\uparrow$}} 
& \multicolumn{3}{c}{\textbf{Consistency (\%) $\uparrow$}} & \multicolumn{2}{c}{\textbf{Accuracy (\%) $\uparrow$}} \\
\cmidrule(lr){4-6} \cmidrule(lr){7-8} \cmidrule(lr){9-11} \cmidrule(lr){12-13}
& & & \textbf{FCR} & \textbf{EEA} & \textbf{EPC} & \textbf{UAR} & \textbf{WAR} 
& \textbf{FCR} & \textbf{EEA} & \textbf{EPC} & \textbf{UAR} & \textbf{WAR} \\
\hline
\rowcolor{gray!15}
\multicolumn{13}{c}{\textit{Non-Reasoning Models}} \\
Former-DFER~\cite{zhao2021former} & \xmark & - & -- & -- & -- & -- & -- & -- & -- & -- & 53.69 & 65.70 \\
MAE-DFER~\cite{sun2023mae} & \xmark & - & -- & -- & -- & 41.62 & 54.31 & -- & -- & -- & 63.41 & 74.43 \\
S2D~\cite{chen2024static} & \xmark & - & -- & -- & -- & 41.86 & 57.37 & -- & -- & -- & 61.82 & 76.03 \\
\hdashline
Emotion-LLaMA~\cite{cheng2024emotion} & \cmark & 7B & -- & -- & -- & -- & -- & -- & -- & -- & 64.21 & 77.06 \\
FaVChat~\cite{zhao2025favchat} & \cmark & 7B & -- & -- & -- & 49.07 & 60.97 & -- & -- & -- & 68.17 & 77.88 \\
Omni-Emotion~\cite{yang2025omni} & \cmark & 7B & -- & -- & -- & 53.81 & 64.23 & -- & -- & -- & 68.80 & 78.35 \\
HumanOmni~\cite{zhao2025humanomni} & \cmark & 7B & -- & -- & -- & 52.94 & 68.40 & -- & -- & -- & 74.86 & 82.48 \\
\hline
\rowcolor{gray!15}
\multicolumn{13}{c}{\textit{Reasoning-Generating Models}} \\
R1-Omni$^{\dagger}$~\cite{zhao2025r1} & \cmark & 7B & 40.22 & 44.97 & 55.19 & 47.27 & 65.36 & 52.01 & 55.39 & 60.70 & 69.69 & 76.67 \\
ERV~\cite{rha2025erv} & \cmark & 7B & 50.98 & 54.70 & 73.06 & 47.01 & 65.19 & 62.06 & 65.50 & 73.53 & 68.88 & 75.81 \\
\textbf{MIGR (Ours)} & \cmark & 7B & 55.30 & 57.65 & 84.37 & 44.82 & 62.46 & 68.48 & 70.06 & 88.95 & 67.78 & 73.93 \\
\bottomrule
\end{tabular}
}
\vspace{-0.3cm}
\label{tab:1}
\end{table*}

\subsection{Implementation Details}
\textbf{Pre-processing.} For video input, we resize each frame to $384 \times 384$ and uniformly sample 8 frames per clip. For audio, we use 16 kHz waveforms and convert them into 128-channel mel-spectrograms using a window size of 25 ms and a hop size of 10 ms.

\noindent \textbf{Architecture.} Following the design of R1-Omni~\cite{zhao2025r1}, our framework consists of SigLIP~\cite{zhai2023sigmoid} as the vision encoder, Whisper-large-v3~\cite{radford2023robust} as the audio encoder, and BERT~\cite{devlin2019bert} as the text encoder. Each modality output is projected into the LLM embedding space through its corresponding visual and audio projectors, implemented as two linear layers to match the dimensionality of the LLM representation. For reference, a 1-second audio clip yields approximately 50 audio features, while a single video frame produces 729 visual features. The LLM backbone is Qwen2.5-7B~\cite{yang2025qwen3}, and all model weights are initialized from the HumanOmni~\cite{zhao2025humanomni} checkpoint to preserve pretrained multimodal representations.

\noindent \textbf{Training and Evaluation.} MIGR consists of two training stages: the SFT stage and the GRPO stage. In the SFT stage, we construct an additional emotional reasoning dataset using the MAFW and DFEW training datasets. Based on the ERV~\cite{rha2025erv}, we align samples where the emotion represented by the reasoning output matches the target emotion. AU (Action Unit) information is extracted using the OpenFace toolkit, and AU sets are organized following the dataset construction pipeline of Emotion-LLaMA~\cite{cheng2024emotion}. Using the Emotion–AU Table for emotion alignment verification, we incorporate 184 samples from MAFW and 253 samples from the DFEW training set as additional reasoning data. Training is conducted for 5 epochs with a cosine scheduler, a warmup ratio of 0.03, a learning rate of 2e-5, and a batch size of 32, using 8~$\times$~NVIDIA~A100 GPUs. In the GRPO stage, the gradient accumulation step is set to 2, the local batch size to 1, and the generation number (G) to 4, with a learning rate of 1e-6, and a cosine decay scheduler. The training is performed for 1 epoch, and evaluation is conducted with a temperature of 0.3.

\section{Experimental Results}
\subsection{Main Result}
\subsubsection{Comparison with the state-of-the-art methods}
To assess the effectiveness of the proposed framework, we compare MIGR with existing emotion understanding methods on the MAFW and DFEW datasets (Table~\ref{tab:1}). We categorize previous methods into two groups: (i) \textit{Non-Reasoning Models} that directly predict emotion labels, and (ii) \textit{Reasoning-Generating Models} that produce textual explanations before making a prediction.

We first compare MIGR with other reasoning-generating models in terms of consistency metrics to validate that the reasoning texts generated by our model are emotionally coherent. As shown in Table~\ref{tab:1}, MIGR consistently outperforms ERV on all three consistency measures across both datasets. On MAFW, MIGR improves over ERV by +4.32, +2.95, and +11.31 points in FCR, EEA, and EPC, respectively, achieving 55.30\% FCR, 57.65\% EEA, and 84.37\% EPC. Similarly, on DFEW, MIGR attains 68.48\% FCR, 70.06\% EEA, and 88.95\% EPC, surpassing ERV by +6.42, +4.56, and +15.42 points. We then compare recognition accuracy between non-reasoning models and reasoning-generating models. Among non-reasoning approaches, HumanOmni achieves the strongest performance with 68.40\% WAR on MAFW and 82.48\% WAR on DFEW, establishing a strong accuracy-oriented baseline. In contrast, all reasoning-generating models, including MIGR and ERV, exhibit slightly lower recognition accuracy than HumanOmni, suggesting that the explicit reasoning generation process can weaken the direct answer-prediction capability. When comparing MIGR with reasoning models, our framework attains 62.46\% WAR on MAFW and 73.93\% WAR on DFEW, which is somewhat lower than the best reasoning counterparts, but accompanied by substantially higher consistency scores.

\subsubsection{Analysis of Reasoning-Answer Inconsistency}
To validate whether the improvement in emotional coherence of MIGR stems from more faithful reasoning, we further examine cases where the reasoning explanation contradicts the target emotion while the predicted answer remains correct. To this end, we utilize two indicators: (i) the proportion of all samples where reasoning is incorrect but the answer is correct (R$\times$ / A$\checkmark$), and (ii) the proportion of such reasoning errors among correctly predicted samples.

As shown in the Table~\ref{tab:2}, MIGR reduces reasoning–answer inconsistency compared to reasoning-generating models, R1-Omni, and ERV across both datasets. On the DFEW dataset, the percentage of inconsistent cases drops from 24.66\% (R1-Omni) and 13.75\% (ERV) to 5.45\% with MIGR, while the relative inconsistency among correct predictions decreases from 32.2\% to 7.37\%. A similar trend is observed on MAFW, with reductions from 25.14\% to 7.16\% (overall) and 38.5\% to 11.5\% (among correct samples). These findings suggest that MIGR not only generates explanations that are more consistent with predicted emotions but also effectively mitigates spurious reasoning that leads to correct answers for the wrong reasons.

\begin{table}[t]
\renewcommand{\arraystretch}{1.1}
\renewcommand{\tabcolsep}{2mm}
\centering
\caption{ Reasoning–answer inconsistency analysis on the MAFW and DFEW datasets. 
R$\times$/A$\checkmark$ represents cases where the reasoning is incorrect while the predicted emotion matches the ground truth.}
\resizebox{0.999\linewidth}{!}{
\begin{tabular}{lcccc}
\toprule
\multirow{3.5}{*}{\textbf{Model}} 
 & \multicolumn{4}{c}{\textbf{R$\times$ / A$\checkmark$ (\%) $\downarrow$}} \\
\cmidrule(lr){2-5}
 & \multicolumn{2}{c}{\textbf{DFEW}} & \multicolumn{2}{c}{\textbf{MAFW}} \\
\cmidrule(lr){2-3} \cmidrule(lr){4-5}
 & \makecell{\textbf{All Samples}} 
 & \makecell{\textbf{Among Correct}} 
 & \makecell{\textbf{All Samples}} 
 & \makecell{\textbf{Among Correct}} \\
\midrule
R1-Omni~\cite{zhao2025r1} & 24.66 & 32.20 & 25.14 & 38.50 \\
ERV~\cite{rha2025erv} & 13.75 & 18.10 & 14.21 & 21.80 \\
\textbf{MIGR (Ours)} & \textbf{5.45} & \textbf{7.37} & \textbf{7.16} & \textbf{11.50} \\
\bottomrule
\end{tabular}
}
\label{tab:2}
\end{table}

\begin{table}[t]
\renewcommand{\arraystretch}{1.1}
\renewcommand{\tabcolsep}{2mm}
\centering
\caption{Emotion-wise accuracy comparison on the DFEW dataset. We report per-emotion recognition accuracy along with overall UAR and WAR. A dagger ($\dagger$) denotes our re-implementation using identical training data. }
\resizebox{0.999\linewidth}{!}{
\begin{tabular}{lccccccccc}
\toprule
\textbf{Method} & \textbf{Hap} & \textbf{Sad} & \textbf{Neu} & \textbf{Ang} & \textbf{Sur} & \textbf{Dis} & \textbf{Fea} & \textbf{UAR} & \textbf{WAR} \\
\midrule
Emotion-LLaMA~\cite{cheng2024emotion} & 93.05 & 79.42 & 72.47 & 84.14 & 72.79 & 3.45 & 44.20 & 64.21 & 77.06 \\
R1-Omni$^{\dagger}$~\cite{zhao2025r1} & 90.59 & 78.04 & 62.29 & 85.71 & 73.04 & 27.59 & 70.56 & 69.69 & 76.67 \\
ERV~\cite{rha2025erv} & 88.14 & 76.72 & 62.85 & 84.56 & 78.84 & 31.03 & 60.00 & 68.88 & 75.81 \\
\textbf{MIGR (Ours)} & 86.50 & 79.10 & 60.04 & 82.03 & 66.89 & 31.03 & 68.89 & 67.78 & 73.93 \\
\bottomrule
\end{tabular}
}
\label{tab:3}
\end{table}

\subsubsection{Emotion-wise Performance Analysis}
To investigate the underlying cause of the reduced accuracy observed in Table~\ref{tab:1} compared to existing methods, we conduct an emotion-wise performance analysis on the DFEW dataset—where Hap (Happiness), Sad (Sadness), Neu (Neutral), Ang (Anger), Sur (Surprise), Dis (Disgust), and Fea (Fear) are the seven evaluation categories—as presented in Table~\ref{tab:3}.

In this experiment, we identify two key findings. First, both reasoning-generating models, ERV and MIGR demonstrate substantially stronger performance than the non-reasoning model (Emotion-LLaMA~\cite{cheng2024emotion}) on the challenging Disgust and Fear categories. This suggests that incorporating explicit reasoning allows models to better capture subtle or ambiguous emotional cues, particularly when predicting emotions such as Fear and Surprise. Second, although MIGR achieves accuracy levels comparable to other reasoning-based approaches and maintains stable performance across most of the remaining emotion classes, it exhibits a pronounced degradation on the Surprise category compared with ERV, with accuracy dropping from 78.84\% to 66.89\%. To more precisely understand this degradation on Surprise, we further analyze the prediction errors of MIGR in comparison with ERV, focusing on the confusion patterns for Surprise and identifying which emotion categories it is most frequently misclassified into.

\begin{table}[t]
\renewcommand{\arraystretch}{1.1}
\renewcommand{\tabcolsep}{2mm}
\centering
\caption{Analysis of the 47 DFEW samples where ERV correctly predicts Surprise but MIGR does not. The table shows MIGR’s misclassification distribution and the frequency with which Surprise appears among the Top-2 reasoning-level predictions.}
\scriptsize
\resizebox{0.98\linewidth}{!}{
\begin{tabular}{lcccccc|c}
\toprule
\multicolumn{8}{c}{\textbf{Samples where ERV is correct (Surprise) but MIGR is incorrect} ($N=47$)} \\
\hline
\rowcolor{gray!15}
\multicolumn{7}{c|}{\textit{MIGR predicted label distribution}} 
& \textit{Surprise Recall (Top-2)} \\
\midrule
\textbf{Predicted label} & Hap & Neu & Ang & Dis & Fea & Unk & Sur \\
\textbf{\# samples}      & 2   & 18  & 3   & 1   & 21  & 2   & 39 \\
\textbf{Ratio (\%)}      & 4.3 & 38.3 & 6.4 & 2.1 & 44.7 & 4.3 & 83.0 \\
\bottomrule
\end{tabular}
}
\label{tab:4}
\end{table}

\begin{table}[t]
\centering
\caption{ Ablation study of the MIGR framework on the DFEW dataset. Stage~1 evaluates the effect of augmentation and reasoning reordering during SFT, and Stage~2 examines the impact of the two MI-guided GRPO rewards.}
\scriptsize
\resizebox{0.95\linewidth}{!}{
\begin{tabular}{lccccc}
\toprule
\multirow{3}{*}{\textbf{Method}} 
& \multicolumn{3}{c}{\textbf{Consistency (\%)$\uparrow$}}
& \multicolumn{2}{c}{\textbf{Accuracy (\%)$\uparrow$}} \\
\cmidrule(lr){2-4} \cmidrule(lr){5-6}
& \textbf{FCR} & \textbf{EEA} & \textbf{EPC} & \textbf{UAR} & \textbf{WAR} \\
\hline
\rowcolor{gray!15}
\multicolumn{6}{c}{\textit{Stage 1: Supervised Fine-Tuning (SFT)}} \\
\midrule
\textbf{Baseline (Used-only EMER)} & 42.21 & 44.86 & 57.41 & 39.16 & 48.33 \\
\quad\textbf{+ Data Augmentation} & 58.99 & 59.97 & 85.19 & 54.60 & 62.80 \\
\quad\quad\textbf{+ Training Data Reordering} & 59.72 & 61.13 & 84.80 & 56.04 & 64.77 \\
\hline
\rowcolor{gray!15}
\multicolumn{6}{c}{\textit{Stage 2: SFT followed by GRPO (init: Stage 1 final model)}} \\
\midrule
\quad\textbf{+ Modality-Aligned Order Reward} 
& 64.64 & 66.14 & 82.92 & 68.18 & 74.91 \\
\quad\textbf{+ Modality-Grounded Reasoning Reward} 
& 67.25 & 68.28 & 88.27 & 67.24 & 73.84 \\
\quad\quad\textbf{+ Both Rewards (MIGR)} 
& 68.48 & 70.06 & 88.95 & 67.78 & 73.93 \\
\bottomrule
\end{tabular}
}
\label{tab:5}
\end{table}

\begin{figure*}[t]
    \centering
    \includegraphics[width=0.9\textwidth]{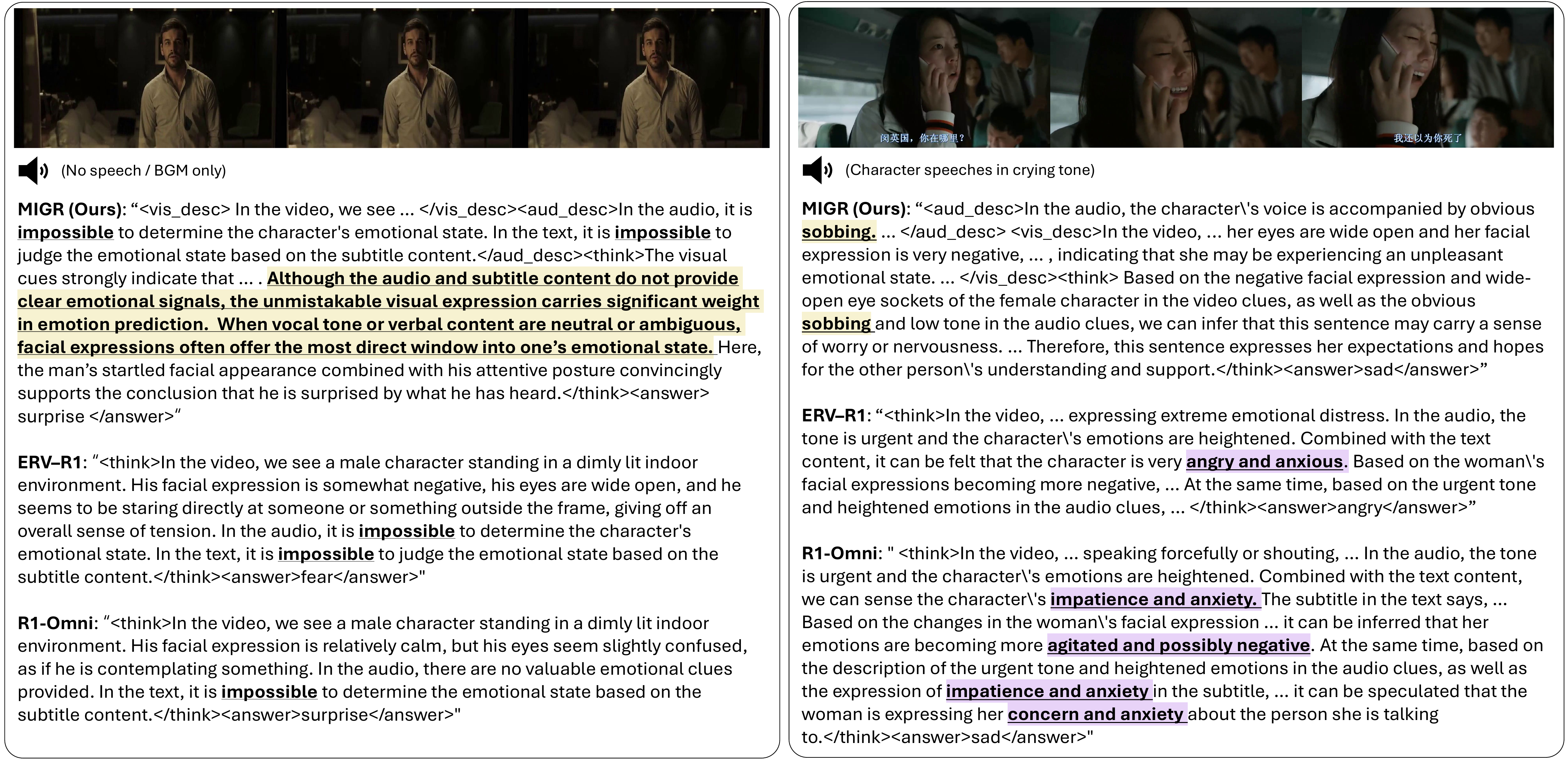}
    \caption{Qualitative comparisons of emotion reasoning. (Left) For a speechless sample, baseline models fail to infer the emotion due to missing audio cues, whereas MIGR correctly focuses on the visual modality and integrates it within the final \texttt{<think>} step to produce a coherent conclusion. (Right) In an audio-dominant sample, MIGR first identifies the key audio cue (“sobbing”) and leverages it to interpret ambiguous visual information, leading to an accurate prediction. In contrast, other models misinterpret the visual cue (frowning) as anger and produce incorrect reasoning.}
    \label{fig:4}
    \vspace{-0.3cm}
\end{figure*}
\subsubsection{Analyzing Performance Degradation on the Surprise Category}
To analyze why Surprise shows particularly lower performance, we examine 47 samples on which ERV correctly predicts Surprise but MIGR fails. As shown in Table~\ref{tab:4}, MIGR predominantly misclassifies these samples as Neutral or Fear. This pattern aligns with the well-known perceptual similarity between Fear and Surprise, which often share facial cues such as widened eyes or raised eyebrows, making them challenging to distinguish even for humans. To further examine whether MIGR truly “loses’’ the Surprise signal, we analyze the generated reasoning using a closed-source LLM. Interestingly, although MIGR’s final predictions are incorrect, Surprise still appears within the Top-2 reasoning-level emotion predictions for 83\% of the samples. This indicates that MIGR retains Surprise-related evidence in its reasoning but ultimately assigns the final label to a neighboring emotion category. Overall, these findings suggest that MIGR’s reduced Surprise accuracy may stem from the inherent ambiguity between Surprise and Fear, indicating that this aspect has room for further improvement.

\subsection{Ablation study}
To better understand the contribution of each component in MIGR, we conduct a stage-wise ablation analysis on the DFEW dataset, as summarized in Table~\ref{tab:5}. Our ablation follows the two-stage training pipeline described in Section~3 and evaluates how successive additions in both SFT and GRPO stages influence emotional consistency and classification accuracy.

We begin with a baseline model trained solely on the EMER dataset, which provides high-quality but limited supervision. Introducing emotion-consistent data augmentation notably improves all consistency metrics, confirming the importance of supplying emotionally reliable samples during early-stage training. Subsequently, applying MI-based reasoning reordering yields additional gains by guiding the model to initiate its reasoning from the emotion-dominant modality.
This validates our hypothesis that organizing reasoning sequences around MI helps mitigate visually biased or text-driven drift in the initial reasoning stage.

In the second stage, we examine how each proposed reward contributes to reasoning stability and coherence. The Modality-Aligned Order Reward encourages the model to begin reasoning from the modality prioritized by MI, restoring the modality-first structure learned during SFT. The Modality-Grounded Reasoning Reward further strengthens the emotional validity of modality-specific reasoning, promoting causal and semantically consistent evidence aligned with the target emotion. When both rewards are applied together, we observe the highest consistency across all metrics, demonstrating that the two rewards provide complementary benefits.

\subsection{Qualitative Results}
Figure 4 presents qualitative comparisons illustrating how MIGR utilizes modality-aligned reasoning to make more reliable emotion predictions. In the first example, the input contains no speech, making audio-based evidence unavailable. Baseline models, ERV, and R1-Omni, therefore fail to describe the emotional state because they are confused by audio or text cues. In contrast, MIGR correctly recognizes that the visual modality is dominant and structures its reasoning accordingly: the model begins with $<vis\_desc>$, identifies key facial cues, and then consolidates the interpretation in the $<think>$ step, leading to a coherent and accurate conclusion. The second example highlights the benefit of modality-grounded reasoning in audio-dominant scenarios. MIGR initiates its reasoning with the audio modality and identifies sobbing, a strong indicator of sadness. This early recognition enables the model to reinterpret the negative visual cues in the context of the audio evidence, resulting in the correct prediction. Baseline models, however, over-rely on visual information and incorrectly interpret the frowning expression as anger, leading to faulty reasoning.

\section{Conclusion}
In this work, we aimed to improve the reliability of reasoning-based multimodal emotion understanding by addressing the early-stage reasoning drift commonly observed in MLLMs. To this end, we introduced MI and proposed MIGR, a training framework designed to initiate and maintain reasoning from the modality most relevant to the target emotion. Our two-stage approach ensures that both the supervised and reinforcement learning processes are aligned with the emotion-dominant modality, leading to explanations that are more coherent, emotionally grounded, and causally meaningful. Experiments on the MAFW and DFEW benchmark show that MIGR greatly reduces emotionally inconsistent reasoning. Overall, MIGR moves toward more trustworthy and interpretable multimodal emotional reasoning, highlighting the value of incorporating modality dominance into both data organization and optimization.

{
    \small
    \bibliographystyle{unsrt}
    \bibliography{main}
}

\clearpage
\maketitlesupplementary

\section{Experimental Setup Details}
\subsection{Datasets}
\label{appendix:appendix7.1}
\noindent \textbf{EMER}~\cite{lian2023EMER} is a small, high-quality emotional video dataset containing 332 samples, each manually verified by humans for both reasoning and answer correctness. It is primarily used during the cold-start SFT stage of MLLM training to initialize the model’s reasoning capability. This dataset contains five emotions: angry, sad, surprise, worried, and happy, and provides corresponding textual descriptions for each sample.

\noindent \textbf{DFEW}~\cite{jiang2020dfew} is a large-scale facial expression video dataset that contains 9,362 training samples and covers seven basic emotions: angry, happy, surprise, disgust, sad, fear and neutral. The dataset additionally provides 2,342 test samples for evaluation.

\noindent \textbf{MAFW}~\cite{liu2022mafw} is a multimodal emotion dataset that contains 7,341 training samples. This dataset includes seven basic emotions as well as four compound categories, namely anxiety, contempt, disappointment and helplessness. It additionally provides 1,831 test samples.

\subsection{Evaluation metrics}
\label{appendix:appendix7.2}
To assess the accuracy of emotion recognition, we adopt two commonly used metrics by following prior works~\cite{cheng2024emotion, zhao2025favchat}: (i) Unweighted Average Recall (UAR), which computes the average recall across all emotion categories without considering their frequencies; and (ii) Weighted Average Recall (WAR), which weights each class by its occurrence proportion to account for class imbalance. Together, these two metrics provide a balanced evaluation of the model’s performance on both frequent and infrequent emotion categories. Formally, let $C$ denote the total number of emotion classes, $N_i$ the number of samples in the $i$-th class, and $TP_i$ the number of correctly predicted samples for class $i$. WAR and UAR are defined as follows:
\begin{align}
    \text{WAR} &= \frac{\sum_{i=1}^{C} \mathrm{TP}_i}{\sum_{i=1}^{C} N_i} \label{eq:war} \\[1em]  
    \text{UAR} &= \frac{1}{C} \sum_{i=1}^{C} \frac{\mathrm{TP}_i}{N_i} \label{eq:uar}
\end{align}

To evaluate the emotional coherence of generated explanations, we leverage three complementary metrics by following~\cite{rha2025erv}: (i) Explanation Emotion Accuracy (EEA), which measures how accurately the emotion expressed in the explanation aligns with the ground-truth emotion label; (ii) Explanation–Prediction Consistency (EPC), which quantifies the degree of consistency between the emotion reflected in the explanation and the model’s predicted emotion; and (iii) Faithful Consistency Rate (FCR), which assesses whether the explanation, the predicted emotion, and the target emotion are mutually consistent. 

In our experiments, we utilize GPT-4.1-mini to infer the emotion expressed in the generated explanation. Given a total of $S$ samples, for the $i$-th sample, let $y_i$ denote the ground-truth emotion, $\hat{y}_i$ the model's predicted emotion label, and $e_i$ the emotion derived from the reasoning text. The metrics are defined as:

\begin{align}
    \text{EEA} &= \frac{1}{S} \sum_{i=1}^{S} \mathbbm{1}[e_i = y_i], \label{eq:eea} \\
    \text{EPC} &= \frac{1}{S} \sum_{i=1}^{S} \mathbbm{1}[e_i = \hat{y}_i], \label{eq:epc} \\
    \text{FCR} &= \frac{1}{S} \sum_{i=1}^{S} \mathbbm{1}[e_i = y_i \land \hat{y}_i = y_i]. \label{eq:fcr}
\end{align}

\section{Qualitative Analysis}
\subsection{Analysis of Attention Distribution}
\begin{figure*}[t!]
    \centering
    \includegraphics[width=0.9\textwidth]{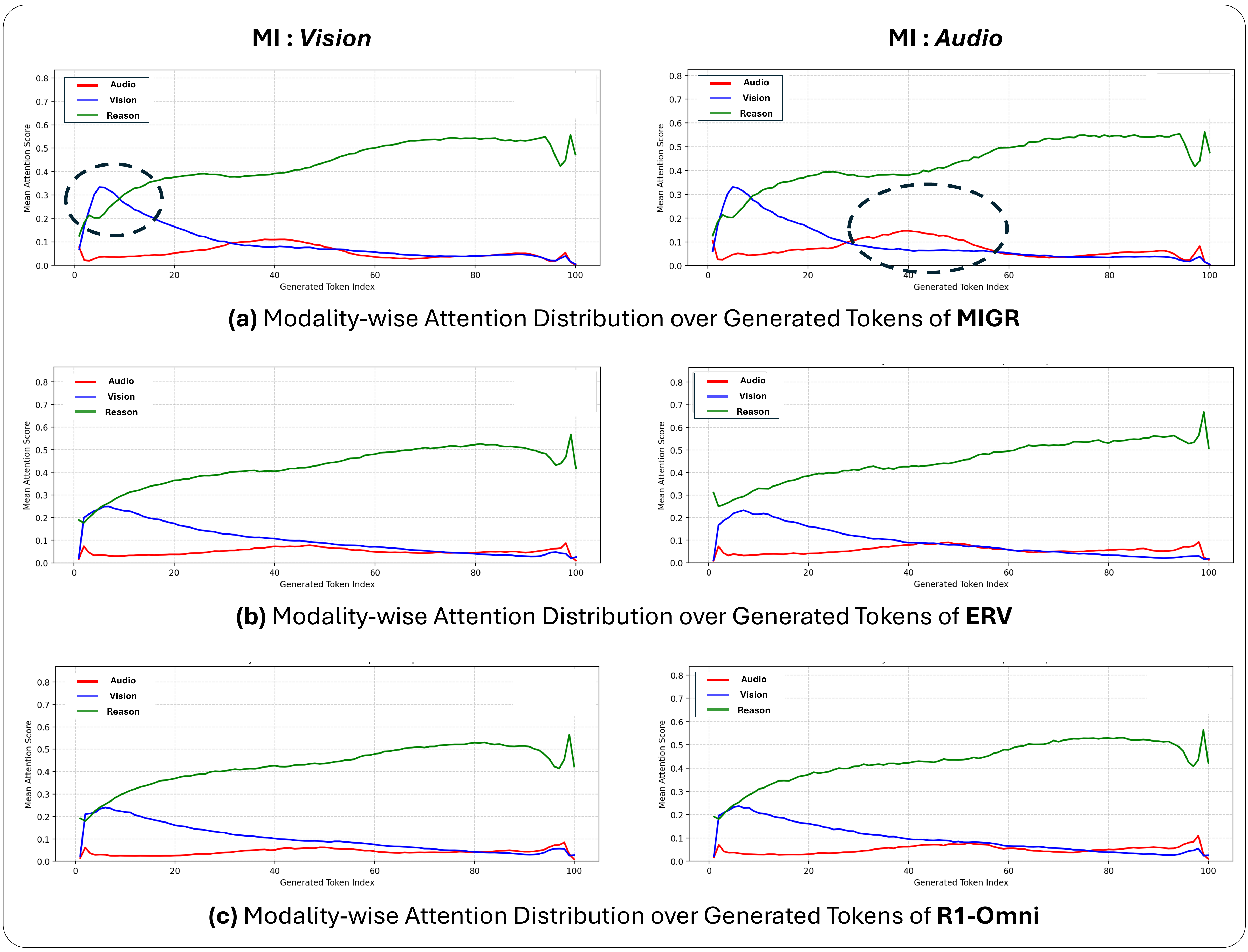}
    \caption{ Comparison of modality-wise attention distributions over generated tokens among MIGR, ERV, and R1-Omni.}
    \label{fig:5}
    \vspace{-0.3cm}
\end{figure*}
To investigate the source of the improvements in consistency metrics, we analyze the attention distribution of the models during the reasoning process. Specifically, we calculate the relative attention scores allocated to visual tokens, audio tokens, and generated text tokens (excluding instruction prompts) at each step of token generation. The analysis was conducted on the DFEW test set, categorized into two subsets based on the MI. To account for varying lengths of generated reasoning across samples, we normalized the generation steps to a fixed length of 100. For the model consisting of 28 layers, we specifically examine the 21st layer, with attention scores averaged across all heads. Figure~\ref{fig:5} illustrates the modality-wise attention distribution over generated tokens for MIGR, ERV, and R1-Omni.

Our analysis reveals two key observations that distinguish MIGR from existing baselines:

\noindent \textbf{Enhanced Grounding on Multimodal Evidence} As shown in Figure~\ref{fig:5}, a prominent difference is that MIGR demonstrates significantly higher peak attention scores for both visual and audio modalities compared to the baseline models, ERV and R1-Omni. This indicates that MIGR is capable of attending to multimodal evidence with greater intensity, thereby reducing reliance solely on language priors. This observation aligns with the qualitative results in Figure~\ref{fig:6}, where baseline models often fail to capture audio cues or generate hallucinations (marked in gray), whereas MIGR produces reasoning well-grounded in the input data.

\noindent \textbf{Dynamic Attention Shift based on Modality Importance} 
Additionally, MIGR exhibits a distinct attention pattern that adapts to the dominant modality. As highlighted by the dashed circles in Figure~\ref{fig:5}(a), when the emotion-dominant modality is Vision ($MI: Vision$), the model's attention to visual tokens is explicitly heightened. Conversely, when the Audio modality is dominant ($MI: Audio$), there is a clear surge in attention toward audio tokens. This contrasts with the baselines, which tend to exhibit a static attention bias (often towards vision or text) regardless of which modality holds the critical emotional information. This demonstrates that MIGR successfully focuses on the informative modality identified by the MI, thereby contributing to the generation of more reliable and emotion-grounded reasoning.

\subsection{Additional Qualitative Results}
\begin{figure*}[t!]
    \centering
    \includegraphics[width=0.9\textwidth]{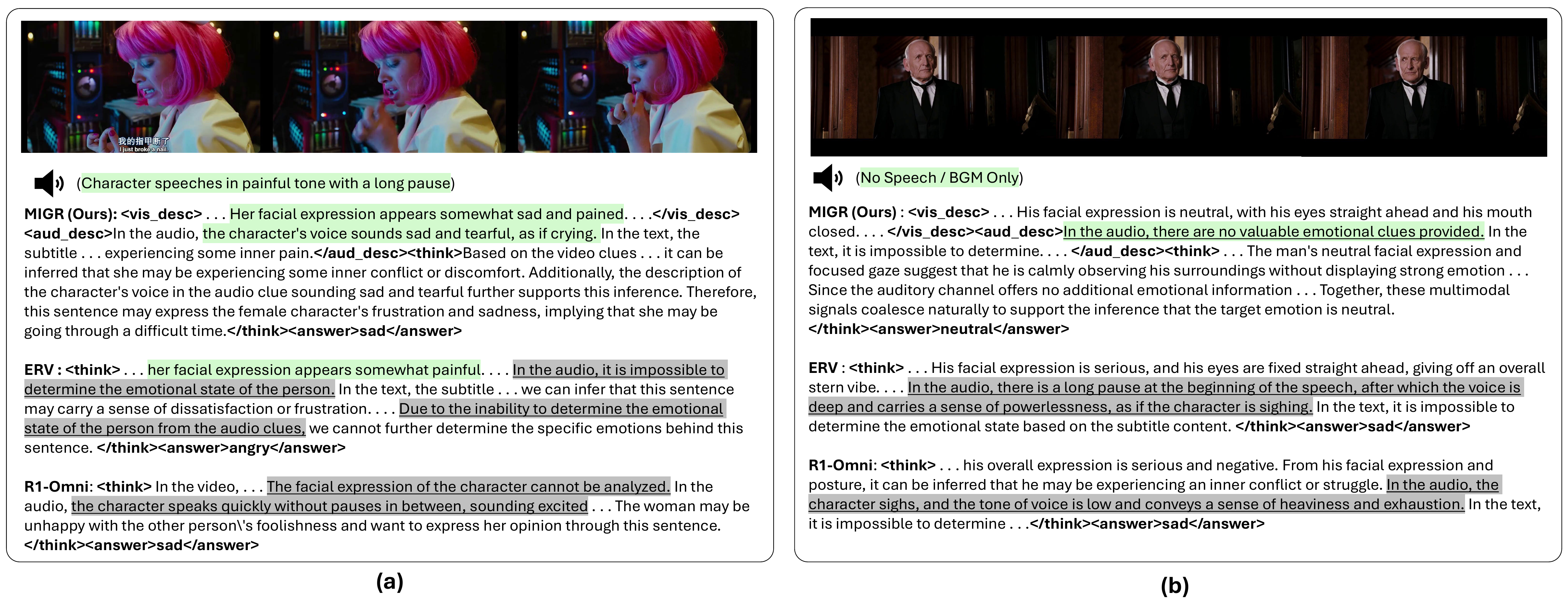}
    \caption{ Qualitative results comparing MIGR, ERV, and R1-Omni. Samples (a) and (b) demonstrate MIGR's capability to precisely reason about auditory emotional cues. }
    \label{fig:6}
    \vspace{-0.3cm}
\end{figure*}
\begin{figure*}[t!]
    \centering
    \includegraphics[width=0.9\textwidth]{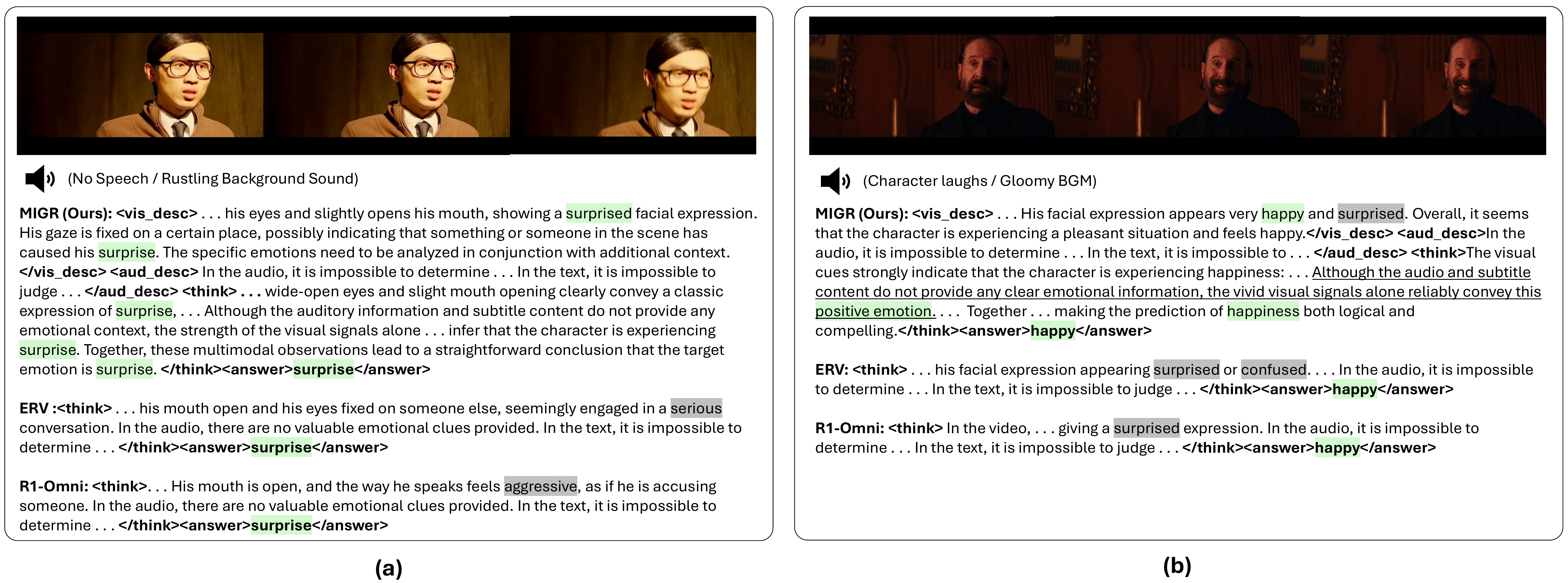}
    \caption{ Qualitative results comparing MIGR, ERV, and R1-Omni. Samples (a) and (b) demonstrate MIGR's capability to consistently align its reasoning with the predicted emotional state. }
    \label{fig:7}
    \vspace{-0.3cm}
\end{figure*}
Figures~\ref{fig:6} and~\ref{fig:7} present concrete examples that further illustrate the distinct advantages of our model compared to baseline methods.

\noindent \textbf{Precise Capture of Auditory Emotional Cues} 
Figure~\ref{fig:6} demonstrates the difference in focusing capability regarding auditory cues. In case (a), despite the presence of the protagonist's clear painful speech, baseline models fail to detect it or provide reasoning with contradictory emotions, such as ``excited.'' Conversely, in case (b), which lacks explicit emotional speech, baseline models tend to hallucinate audio descriptions corresponding to the serious visual cues. In contrast, MIGR accurately captures the ``sad and tearful'' audio signals in (a) and correctly infers the absence of emotion-related information in the audio of (b), demonstrating superior audio-visual discrimination.

\noindent \textbf{Alignment between Reasoning and Prediction} 
Figure~\ref{fig:7} highlights the improved consistency performance of our approach. In (a), MIGR maintains the ``surprise'' emotion consistently from the initial reasoning phase to the final prediction. In (b), while MIGR initially considers both ``happy'' and ``surprise,'' it successfully refines its judgment through a progressive reasoning process, converging to ``happy'' as the final prediction. On the other hand, baseline models exhibit reasoning processes that are inconsistent with their final predicted emotions. These qualitative results provide concrete examples that corroborate the consistency analysis presented in Table~\ref{tab:2}.

\begin{figure*}[h]
    \centering
    \includegraphics[width=0.9\textwidth]{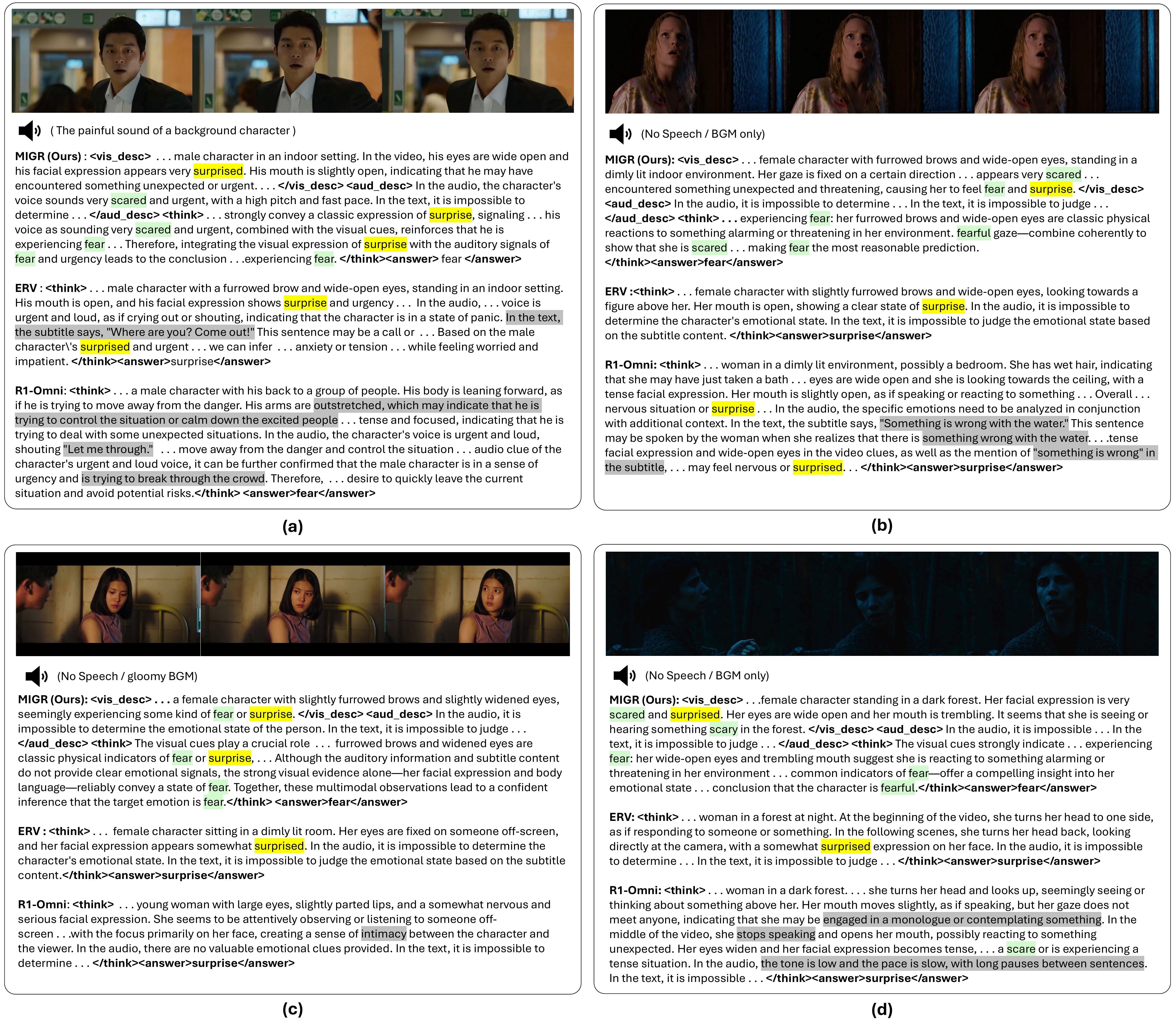}
    \caption{ Qualitative comparison of samples with Surprise as the ground truth. While MIGR misclassifies the final label as Fear, its reasoning process accurately captures the compound emotional state of both Surprise and Fear. In contrast, ERV focuses solely on a single emotion in its reasoning, and R1-Omni exhibits hallucinations by basing its inference on non-existent subtitles. Text highlighted in yellow and green corresponds to reasoning associated with Surprise and Fear, respectively, while gray highlights indicate hallucinations regarding content absent from the input video.}
    \label{fig:8}
    \vspace{-0.3cm}
\end{figure*}
\section{Additional Analysis of Performance Degradation in the Surprise Category}
\subsection{Analysis of Annotation Ambiguity}

\begin{table}[h]
\renewcommand{\arraystretch}{1.2} 
\renewcommand{\tabcolsep}{1.8mm}
\centering
\caption{Analysis of \textbf{concurrent human annotations} for the 47 Surprise samples misclassified by MIGR. Although the ground truth is Surprise, a significant portion of annotators voted for \textbf{Neutral} and \textbf{Fear}, indicating the inherent ambiguity and compound nature of these samples.}
\scriptsize
\resizebox{0.98\linewidth}{!}{
\begin{tabular}{lcccccc|c}
\toprule
\multirow{2}{*}{\textbf{Metric}} & \multicolumn{6}{c|}{\textit{Concurrent Emotion Annotation}} & \textit{Total Target} \\
 & Hap & \textbf{Neu} & Ang & Dis & \textbf{Fea} & Sad & \textit{Samples} \\ 
\midrule
\textbf{\# Samples} & 0 & \textbf{35} & 9 & 4 & \textbf{23} & 5 & 47 \\
\textbf{Ratio (\%)} & 0.0 & \textbf{74.5} & 19.1 & 8.5 & \textbf{48.9} & 10.6 & 100.0 \\
\bottomrule
\end{tabular}
}
\label{tab:6}
\end{table}
Although the DFEW dataset is utilized as a single-label benchmark, its ground truth generation involves a voting process by 10 trained human annotators. A video is assigned a single emotion label only if it receives a consensus vote of over 60\%. Importantly, other emotions perceived in the video are also recorded as concurrent labels.
To investigate the misclassification of Surprise samples by MIGR, we analyzed the distribution of these concurrent annotations for the 47 samples where MIGR's prediction diverged from the ground truth. As detailed in Table~\ref{tab:6}, we found that a significant portion of these samples received annotator votes for the emotions predicted by MIGR. Specifically, 48.9\% of these samples contained annotations for Fear, and 74.5\% included votes for Neutral. This high overlap between the concurrent human annotations and MIGR's predictions (Table~\ref{tab:4}) suggests that MIGR is not merely making erroneous predictions; rather, it is sensitive to the subtle, multi-faceted nature of emotions that even human annotators perceive differently.

\subsection{Qualitative Analysis of Performance Degradation}
Building on this insight, we revisited the qualitative performance discussed in Sections 5.1.3 and 5.1.4. While a notable portion of Surprise samples were classified as Fear, our analysis shows that MIGR successfully captures the Surprise element within its reasoning process in 83\% of these cases (as shown in Table~\ref{tab:4}).

Consistent with the previous annotation analysis, our qualitative inspection reveals that these misclassified samples often exhibit compound emotions, where facial cues. The generated reasoning text confirms that MIGR accurately recognizes this complexity, explicitly discussing the presence of both Fear and Surprise elements.

In contrast, while the ERV often matched the ground truth label, its reasoning tended to focus exclusively on the Surprise features, failing to capture the co-occurring Fear signals or the emotional ambiguity. Furthermore, we observed critical hallucination issues in baselines. For example, as shown in Figure~\ref{fig:8}(a), in a scene containing only background sound, ERV and R1-Omni generated reasoning based on non-existent dialogue (e.g., ``Where are you? Come Out!"). This indicates a susceptibility to text-bias, leading to fabricated evidence. Conversely, MIGR effectively mitigates such hallucinations and provides robust reasoning by anchoring its output to the actual multimodal evidence, attributed to our Modality-Grounded Reasoning objective.

\end{document}